\pdfoutput=1

\documentclass[11pt]{article}

\usepackage[final]{acl}

\usepackage{times}
\usepackage{latexsym}
\usepackage{booktabs}
\usepackage{multicol}
\usepackage{float}
\usepackage{multirow}
\usepackage{stfloats}

\usepackage[T1]{fontenc}

\usepackage[utf8]{inputenc}

\usepackage{microtype}

\usepackage{inconsolata}

\usepackage{graphicx}

\title{Transferring Extreme Subword Style Using Ngram Model-Based Logit Scaling}

\author{Craig Messner \\
  Center for Digital Humanities \\
  Johns Hopkins University \\
  \texttt{cmessne4@jhu.edu} \\\And
  Tom Lippincott \\
  Center for Digital Humanities \\
  Johns Hopkins University \\
  \texttt{tom.lippincott@jhu.edu} \\}

\begin{document}
\maketitle
\begin{abstract}
We present an ngram model-based logit scaling technique that effectively transfers extreme subword stylistic variation to large language models at inference time. We demonstrate its efficacy by tracking the perplexity of generated text with respect to the ngram interpolated and original versions of an evaluation model. Minimizing the former measure while the latter approaches the perplexity of a text produced by a target author or character lets us select a sufficient degree of adaptation while retaining fluency.
\end{abstract}

\section{Introduction}

Text style transfer (TST) aims to reformulate a source text using the stylistic attributes of a given target text. Authors vary a blend of these attributes to achieve a literary effect, with some modifications being more conspicuous than others. Stylistic modification of subword units like characters or phonemes can prove especially noticeable.\footnote{In this paper, all fundamental units are subwords, as such references to tokens refers to subword tokens produced by subword tokenization methods.} 

One such subword style is orthographic variation, a phenomenon common in forms of "dialect literature" present throughout history but especially popular in the 19th century United States \cite{krapp1925english} \cite{ives1971theory}. These works utilize context and readerly knowledge to render their orthographic innovations legible despite their extreme modification of orthographic norms \cite{sebba2007spelling}.
We present a subword-level ngram-based logit scaling method that effectively transfers this form of extreme style at decoding time. We accomplish this by combining the next-token information derived from a large language model (LLM) with information obtained from ngram models trained on a single-author/character corpus.

Ngram models are quick to train, data-efficient, and interpretable. Training ngram models on small single-author corpora re-purposes them as statistical experts, reflections of the constructions a given author is likely to employ. These qualities are especially useful when transferring style from low frequency or novel sources. LLMs may have little or no information about these styles in their weights, and style-specific corpora may be too small to support finetuning. 

We introduce a scaled interpolation method that combines weighted ngram model predictions with those from pretrained LLMs to generate fluent stories that match the extreme subword style of particular dialect authors and characters. We also demonstrate how to tune and evaluate these transfers using perplexity measures.

\section{Related Work}
Techniques like finetuning on further data \cite{mukherjee-etal-2024-large-language}, prompt editing \cite{luo-etal-2023-prompt} and in-context learning \cite{mai-etal-2023-prefix} have been used to achieve TST. While potentially effective, these avenues require further computation and additional training data. Mechanical interpretation approaches can provoke style at inference time by intervening on model weights \cite{lai-etal-2024-style}. However, this approach requires the target style to be in-distribution and suitably represented in the model. Other recent works have re-evaluated LM approaches previously considered obsolete in the light of computational and theoretical advances. Ngram modeling has been revisited for LM smoothing \cite{malagutti-etal-2024-role} and in "infinite" form as a interpolation component used to complement LLMs \cite{liu2024infini}.

\section{Methodology}
We achieve subword TST by applying an ngram model-derived scaling factor to the output logits of an LLM prior to softmaxing and sampling. Vitally, information from the ngram model must contribute to the next-token probability without warping the LLM's ability to produce fluent text. We ensure both fluency and transfer by scaling the ngram model-provided next token prediction with an additional factor $f$. Given a vocabulary $V$, we calculate the scaling factor using equation~\ref{eq:eq1}

\begin{equation}
    \label{eq:eq1}
    S = -\frac{f}{log(p_n)} n \in \{0...|V|\}  
\end{equation}

Inspired by temperature decoding methods, the addition of the parameter $f$ uniformly increases the scaling factor as $f$ increases, leading to a higher proportion of generation information being derived from the ngram model. This also renders the scaling mixture "tunable". 

\section{Experiments}
\subsection{Setup}

\textbf{Data.} We use two 19th century U.S. fiction corpora sourced from Project Gutenberg (PG) for baseline evaluation and ngram model training. The first consists of the full text of \textit{Uncle Remus: His Songs and His Sayings} by Joel Chandler Harris and all of Peter Finley Dunne's \textit{Mr. Dooley} series of stories. The second corpus consists of dialogue employed by three characters identified as nonstandard speakers. Messner extracted these dialogue sections from PG and attributed them manually. See Appendix~\ref{sec:furthercorp} for more corpus details. We style the former corpus with lower casing and the latter with upper. We wordpiece tokenize both using the model-specific functions supplied by the Transformers library. 

We generate story prompts and establish standard English baseline scores using the \textsc{writingprompts} (WP) dataset \cite{fan-etal-2018-hierarchical}. We sample 50 prompts from the dataset to guide story generation. We modify each original prompt with a brief instruction stem in order to produce the scaled generation prompt set. Additionally, we apply three character/author templates to each prompt in order to produce a control prompt set. The first indicates that the model should act as a storyteller, and includes a description of its era and position. The second adds information about the target author, and the third the target character. See Appendix~\ref{sec:prompts} for more prompt creation details.

\textbf{Models.} For generation of control and scaled texts, we use MistralAi's Mistral-Instruct-7B-v0.2 (Mistral) \cite{jiang2023mistral} and Meta's Llama3.2-3B-Instruct (Llama) \cite{dubey2024llama} pretrained instruction-tuned models. For perplexity evaluation, we use OpenAI's GPT2-large (GPT2) \cite{radford2019language}. We obtain model checkpoints via HuggingFace. Using the wordpieced target texts, we train a set of ngram models, $\{M_n, M_{n-1},..., M_1\}$, with $n=4$. When scaling generated logits, we employ the model set in a backoff configuration. If $M_4$ cannot make a 4gram next token prediction, we use a trigram prediction from $M_3$, and so on. If no model can make a prediction, no scaling is performed. This is essentially a modification of stupid backoff \cite{brants-etal-2007-large}.

\textbf{Evaluation.} We concatenate the tokens produced by each scaled or control generation and estimate their GPT2 perplexity using a sliding context window of 32 tokens with stride of 1. We do the same for the WP test-set baseline and GB target texts. When scoring general model performance, low perplexity is considered preferable. For our purposes, near-equal target ($PPL()$) and generation ($gPPL()$) perplexities indicate successful subword style transfer. We also measure the perplexity of the original texts using the interpolation of GPT2 and each target text's scaled ngram models, $rPPL()$. Combining these two sources of information allows us to select an optimal schedule of scalings for subword style transfer by maximizing $abs(PPL()-gPPL())$ while minimizing $rPPL()$. Intuitively, the first measure acts as an early stopping criterion, while the second measure indicates whether the $gPPL()$ at a given scaling is produced by transfer and not chaotic.

\subsection{Procedure.}
We define a 16-member weight set $W$, where each $w\in W$ is a tuple of the form $\{f_4,f_3,f_2,f_1\}$. Each $f$ is drawn from $\{0,1,2\}$ and used to scale the next token predictions $p$ of its corresponding ngram set model $M_n$ using Equation~\ref{eq:eq1}. $f$ of 0 omits the corresponding model. For example, $w$ of $\{0,0,2,1\}$ applies Equation~\ref{eq:eq1} with $f=2$ and $f=1$ to the bigram and unigram model next-token probabilities respectively. This results in a scaling vector $S$ with length $|V|$. We add $S$ to the logits of the LLM's next token prediction and repeat the process up to a maximum generation length of 256 tokens. We perform $|W|$ of these scaled generations for each prompt in the base set, using a different $w$ each time. We repeat this process over two conditions: decoding greedily and sampling. We calculate $gPPL()$ and $rPPL()$ and then graphically determine the weight set(s) of best fit for a given target character or author by plotting $abs(PPL() - gPPL())$ against $rPPL()$ . \footnote{Code and data for these experiments available at: \url{https://github.com/comp-int-hum/llm-decode-style}} 

\section{Results and Discussion}
\subsection{Baselines and target styles}

The $PPL()$ of the baseline (WP) and variant target texts (GB) greatly differ (Table~\ref{tab:baseline}). The target texts are considerably more perplexing, at least in part due to the modifications they employ at the subword level. Consequently, a $gPPL()$ more similar to the target $PPL()$ than the baseline WP $PPL()$ indicates that style was likely transferred.

\begin{table}[t]
\centering
\begin{tabular}{lrrrrrr}
\toprule
Target & $PPL()$ & N Tokens  \\
\midrule
remus & 106.54 & 82365 \\
dooley & 110.03 & 366037 \\
\midrule
Todd & 49.88 & 12273 \\
Remus & 128.68 & 48217  \\
Julius & 166.51 & 11350 \\
\midrule
WP  & 41.01 & 12456693 \\
\bottomrule
\end{tabular}

\caption{Baseline results. Top section: full texts from GB. Middle section: Dialogue extracted from GB. Bottom section: WP (standard) baselines}
\label{tab:baseline}
\end{table}

\subsection{Generation conditions and model specificity}

Neither scaled LLM produces text with $gPPL()$ approaching its particular target $PPL()$ when greedy decoding is employed.

However, when sampling is employed instead, scaled Mistral produces text with $gPPL()$ closest to those of the target texts. See Appendix~\ref{sec:unsuc} for the numerical results. Llama3.2 consistently falls short of the targets. Differences in pretraining data and instruction-tuning regimes likely explain this performance disparity.

\subsection{Control results}

\begin{table}[h]
\centering
\addtolength{\tabcolsep}{-0.2em}
\begin{tabular}{lrrrrrrr}
\toprule
 Prompt & Remus & Todd & Julius & remus & dooley \\
\midrule
1 & 23.31 & 23.31 & 22.20 & 23.31 & 22.37 \\
2 & 21.47 & 22.51 & 20.15 & 21.47 & 24.04 \\
3 & 41.88 & 18.92 & 19.74 & 41.88 & 30.66 \\
\bottomrule
\end{tabular}

\caption{$gPPL()$ of sampled unscaled Mistral logits for each of the three control prompts}
\label{tab:mistralbase}
\end{table}

Unscaled LLM generation over the control prompts did not result in appropriate $gPPL()$ scores (Table~\ref{tab:mistralbase}) indicating that the extreme elements of style were largely not transferred (see Appendix~\ref{sec:appcs} for a sample generation).

However, unscaled Mistral was able to produce some appropriate subword features when provided with the Remus and dooley versions of the third prompt. Take this sample generated from the Remus version:

\begin{quote}
    Ah, chilren, dis here's a mighty strange tale dat comes to us from de big screen. Leonardo DiCaprio, he was once a fine actor, like a fish swimmin' gracefully in a crystal-clear stream.
\end{quote}

While "chilren, dis" is likely a high-probability generation for Remus, "gracefully in a crystal-clear stream" is likely not. Relying solely on prompt construction to evoke subword style is both fragile and coarse. While there may be some prompt $p$ that is able to evoke further Remus subword style from the model, thereby increasing the generation's $gPPL()$ towards $PPL()$, it is unclear how to construct this prompt. Furthermore, it is not clear that modifying $p$ could in any case elicit subword style for the non-Remus authors/characters. 

\subsection{Scaled generation results}

\begin{figure}[h]
\includegraphics[width=.45\textwidth]{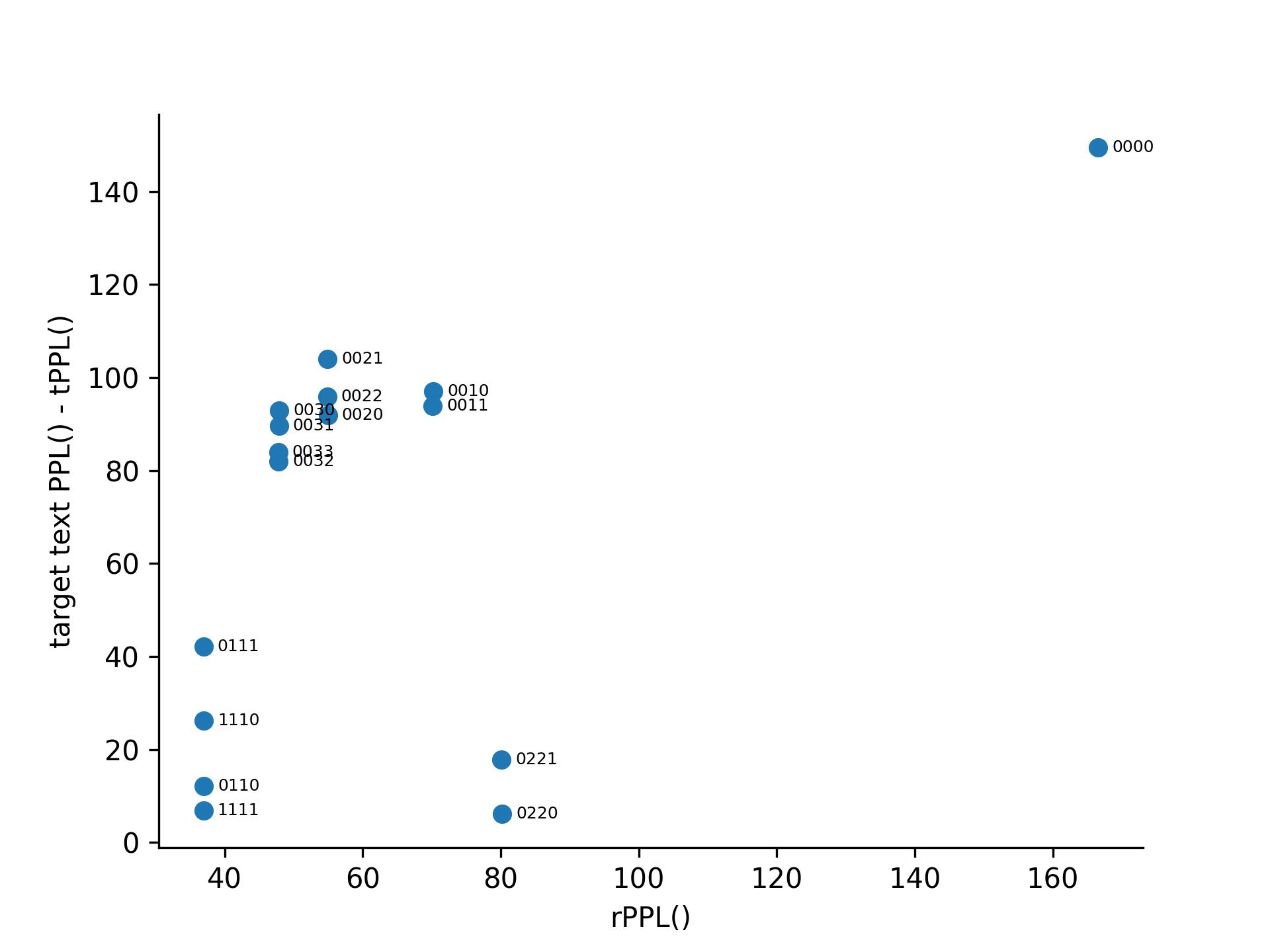}
\caption{Julius $abs(PPL()-gPPL())$ and $rPPL()$. Optimal scalings are located in the bottom-left corner.}
\label{fig:Julius}
\end{figure}

\begin{table*}[t]
\centering
\small
\begin{tabular}{lrrrrrrrrrrrr}
\toprule
 & \multicolumn{2}{c}{Remus} & \multicolumn{2}{c}{Todd} & \multicolumn{2}{c}{Julius} & \multicolumn{2}{c}{remus} & \multicolumn{2}{c}{dooley} \\
 weights & gPPL() & rPPL() & gPPL() & rPPL() & gPPL() & rPPL() & gPPL() & rPPL() & gPPL() & rPPL() \\
\midrule
0111 & 87.55 & 36.94 & 75.14 & 73.97 & 124.46 & 26.74 & \textbf{109.87} & \textbf{50.34} & 125.46 & 229.26 \\
1111 & 86.61 & 36.94 & 79.22 & 73.97 & \textbf{159.66} & \textbf{26.74} & 89.57 & 50.34 & \textbf{117.32} & \textbf{229.26} \\
0110 & 92.08 & 36.99 & 89.15 & 74.04 & \textbf{154.42} & \textbf{26.76} & \textbf{96.95} & \textbf{50.35} & 124.31 & 229.38 \\
1110 & \textbf{108.04} & \textbf{36.99} & 80.32 & 74.04 & \textbf{140.28} & \textbf{26.76} & \textbf{99.95 }& \textbf{50.35} & \textbf{119.28} & \textbf{229.38} \\
0033 & 57.29 & 47.79 & \textbf{50.16} & \textbf{53.51} & 82.56 & 23.98 & 58.43 & 82.88 & 40.70 & 89.68 \\
0032 & 57.21 & 47.82 & \textbf{47.78} & \textbf{53.55} & 84.51 & 23.99 & 52.14 & 82.90 & 42.64 & 89.72 \\
0021 & 55.15 & 54.92 & \textbf{52.41} & \textbf{57.33} & 62.58 & 25.45 & 51.98 & 78.73 & 40.93 & 78.64 \\
\bottomrule
\end{tabular}

\caption{Optimal $gPPL()$ and $rPPL()$ for sampled scaled Mistral logits. Bolded values are the graphically-determined best performers for a target text. Suboptimal scalings are found in Appendix~\ref{sec:misunsuc}}
\label{tab:missamp}
\end{table*}

Given the above, we derive our main results from generations produced by sampling the scaled Mistral distribution (Table~\ref{tab:missamp}). We select the Julius scaled results for further inspection (Figure~\ref{fig:Julius}), and choose a sample generation produced by one of the optimal conditions, [1,1,1,0], to guide further discussion (Table~\ref{tab:juliusgenex}).

\textbf{Scaling successfully mixes information from both models.}
Tokens that begin a proper name or noun are frequently selected even when their corresponding logits were not scaled, implying that their prior probability as conditioned by the story prompt was not eclipsed by information from the ngram model. Proper names and nouns were also frequently completed with their standard continuations, likely due to the low internal entropy of the wordpieces. This includes the names of central story characters, e.g. \textit{Di}+\textit{Cap}+\textit{rio}.

Important functional words like pronouns and conjunctions are frequently selected for grammatically appropriate positions, demonstrating that Mistral's generation was not negatively impacted by the addition of ngram scaling information.

Author-characteristic continuations such as \textit{w}+\textit{’}+\textit{en} instead of \textit{w}+\textit{hen} were selected from the probability distribution, demonstrating that ngram scaling weighted these wordpieces enough to overcome Mistral's preference for the standard form.

\textbf{Scaling produces novel, plausible sequences.} Mistral combined with ngram scaling produces author and character-plausible sequences not present in the target text:

\begin{quote}
    \textit{p}+\textit{us}+\textit{se}+\textit{w}, \textit{rep}+\textit{u}+\textit{’}+\textit{ce}+\textit{ation}
\end{quote}

In effect, such sequences are hypotheses about how a character/author might style particular words that go unused in their actual corpus.

\textbf{Optimal scaling is style specific.} Table ~\ref{tab:missamp} shows that optimal transfer of Todd's "backwoods" style requires a unique set of scaling weights. Her subword style, characterized by minor elisions (e.g. hopin' instead of hoping), is closer to standard American orthography than those used by Julius or Remus, thus increasing the utility of LLM information in the transfer process.

\textbf{Controlling scaling weights allows for generating "degrees" of the targeted style.} Selecting slightly "suboptimal" scaling leads to a smooth interpolation of subword style into the generated text. Appendix~\ref{sec:near} contains an example of this graded interpolation. 

\begin{table}[H]
    \small
    \centering
    \textbf{Sample generation: Julius [1,1,1,0]}
    \begin{tabular}{p{\linewidth}}
      [INST]Write a few sentences based on the following story prompt:  Leonardo DiCaprio in a fit of rage begins to torpedo his own career by deliberately acting poorly and taking on bad films. He finally wins an oscar for starring in Paul Blart: Mall Cop 3. [/INST] \\[2mm]
    
     In \textcolor{blue}{de} \textcolor{orange}{ma} \textcolor{orange}{wn} \textcolor{orange}{in} \textcolor{orange}{'} \textcolor{orange}{glow} of \textcolor{blue}{a} \textcolor{blue}{sm} \textcolor{blue}{old} ering Hollywood career , \textcolor{blue}{Leon} ardo Di Cap rio , \textcolor{blue}{the} erst while \textcolor{blue}{golden} boy \textcolor{blue}{fer} \textcolor{orange}{a} \textcolor{orange}{generation} , \textcolor{blue}{stood} before a \textcolor{blue}{full} \textcolor{blue}{er} \textcolor{orange}{cow} \textcolor{orange}{-} \textcolor{orange}{pe} \textcolor{orange}{as} \textcolor{orange}{.} \textcolor{blue}{He} \textcolor{orange}{w} \textcolor{orange}{uz} \textcolor{orange}{a} \textcolor{orange}{man} \textcolor{orange}{in} \textcolor{orange}{dis} \textcolor{orange}{settlement} \textcolor{orange}{w} \textcolor{orange}{'} \textcolor{orange}{en} \textcolor{orange}{he} \textcolor{orange}{w} \textcolor{orange}{uz} \textcolor{orange}{in} \textcolor{orange}{a} \textcolor{orange}{fit} \textcolor{blue}{ter} \textcolor{orange}{kill} \textcolor{orange}{his} \textcolor{blue}{own} \textcolor{blue}{rep} u \textcolor{blue}{'} \textcolor{orange}{ce} ation \textcolor{blue}{.} \textcolor{blue}{F} \textcolor{blue}{us} \textcolor{orange}{in} \textcolor{blue}{'} \textcolor{orange}{w} \textcolor{orange}{if} a \textcolor{blue}{wr} \textcolor{blue}{ath} ful passion , \textcolor{blue}{he} \textcolor{orange}{hur} led \textcolor{blue}{r} \textcolor{orange}{oun} \textcolor{orange}{'} \textcolor{orange}{reck} \textcolor{blue}{on} \textcolor{orange}{in} \textcolor{blue}{'} \textcolor{orange}{s} \textcolor{orange}{,} \textcolor{orange}{intent} on \textcolor{blue}{self} - \textcolor{blue}{dest} ruction . \textcolor{blue}{<0x0A>} <0x0A> Di Cap rio , \textcolor{blue}{a} \textcolor{orange}{man} \textcolor{orange}{in} \textcolor{orange}{dis} \textcolor{orange}{settlement} \textcolor{orange}{w} \textcolor{orange}{'} \textcolor{orange}{en} \textcolor{orange}{he} \textcolor{orange}{w} \textcolor{orange}{uz} \textcolor{orange}{,} \textcolor{blue}{began} to \textcolor{blue}{p} \textcolor{blue}{us} \textcolor{blue}{se} \textcolor{blue}{w} \textcolor{blue}{projects} that sc \textcolor{blue}{upper} \textcolor{orange}{non} \textcolor{orange}{'} \textcolor{orange}{d} \textcolor{orange}{reason} , \textcolor{blue}{le} \textcolor{blue}{av} in \textcolor{blue}{'} \textcolor{orange}{a} \textcolor{orange}{fl} \textcolor{blue}{uster} \textcolor{orange}{ated} \textcolor{orange}{dat} \textcolor{orange}{we} \textcolor{orange}{fu} \textcolor{orange}{h} \textcolor{orange}{ter} \textcolor{blue}{in} \textcolor{orange}{'} \textcolor{orange}{a} \textcolor{orange}{gh} ast \textcolor{blue}{le} \textcolor{orange}{wid} \textcolor{orange}{'} \textcolor{orange}{im} \textcolor{orange}{.} \textcolor{orange}{He} \textcolor{orange}{w} \textcolor{orange}{uz} \textcolor{orange}{the} ant agon ist of \textcolor{blue}{'} \textcolor{blue}{is} \textcolor{orange}{aw} ' \textcolor{blue}{n} \textcolor{orange}{tales} w \textcolor{blue}{if} a \textcolor{blue}{fierce} , \textcolor{blue}{reck} \textcolor{blue}{on} \textcolor{orange}{in} \textcolor{blue}{'} \textcolor{orange}{g} \textcolor{orange}{lee} , \textcolor{blue}{sign} in \textcolor{blue}{'} \textcolor{orange}{contract} after ill - \textcolor{blue}{con} ceived contract to \textcolor{blue}{appear} ter \textcolor{blue}{ribly} in \textcolor{blue}{films} that ' \textcolor{blue}{d} \textcolor{orange}{long} \textcolor{blue}{leave} dis \textcolor{blue}{realm} of \textcolor{blue}{memory} . \textcolor{blue}{<0x0A>} <0x0A> F ew believed Di Cap rio w \textcolor{blue}{uz} \textcolor{orange}{truly} mad \textcolor{blue}{der} \textcolor{orange}{dan} \textcolor{blue}{a} \textcolor{blue}{h} \textcolor{orange}{atter} , \textcolor{blue}{but} \textcolor{orange}{it} \textcolor{orange}{w} \textcolor{orange}{uz} \textcolor{orange}{a} \textcolor{orange}{certain} ty wid \textcolor{blue}{dis} \textcolor{blue}{actor} ' \textcolor{blue}{s} \textcolor{orange}{met} ic ulous craft sm ans hip he \textcolor{blue}{w} \textcolor{orange}{uz} \textcolor{orange}{m} \textcolor{blue}{ak} \textcolor{blue}{tree} \textcolor{orange}{-} \textcolor{blue}{m} \textcolor{blue}{end} ously bad \textcolor{blue}{deliber} at 
     
    \end{tabular}
    \caption{Generation using the Julius extracted dialogue ngram model, sampled from scaled Mistral distribution. Blue tokens are bigram scaled, orange trigram scaled.}
    \label{tab:juliusgenex}
\end{table}

\section{Conclusions and Further Work}
Our ngram scaling method produces plausible story generations that bear features of the extreme subword style of their target author or character in a compute and data-efficient manner. Further work can be performed to test the method on other forms of subword variation, and to characterize the specific subword features that were transferred relative to the subword tokenization system used by a given LLM and ngram model. 

Additional work could also include increasing the precision of our method for determining scaling optimality, further characterizing a scale of subword-style extremity in order to help determine what forms of style are likely candidates for transfer by this method, and experimenting with hybrid generation across multiple author ngram models.

\section{Limitations}
We currently only apply our approach to authors and characters drawn from 19th century United States literature. Other eras, nationalities, and in particular, languages, may employ subword variations our method cannot transfer. Currently, this method depends on the subword tokenization systems used by pretrained LLMs. The learned boundaries their wordpiecing systems employ could omit some elements of subword style.

\section{Ethical Considerations}
Automating style transfer increases the risk of sophisticated stylistic forgery. However, the type of style transferred in this case is primarily archaic, and typically used for literary, rather than personal, ends, considerably lessening this approach's nefarious utility.

Some of the texts used to test this method are controversial as they can be seen as caricaturing their subjects. These texts also commonly employ offensive terminology. The nature of our method means that these attributes may be expressed at generation time. However, these styles were influential, and thus of literary-historical importance, and should be studied despite these issues. 

\bibliography{anthology.bib, custom.bib}
\clearpage
\appendix
\section{Further Corpus Details}
\label{sec:furthercorp}
Further information about the hand-attributed dialogue corpus:
\begin{enumerate}
    \item Remus: Harris's titular storyteller, as extracted from the full remus text. Part of the "plantation literature" genre. An extreme form of African American English. 
    \item Julius: Julius McAdoo, the storyteller of Charles Chesnutt's \textit{Conjure} tales. Frequently considered to be "anti-plantation literature." An extreme form of African American English.
    \item Todd: Almira Todd, the narrator of Sarah Orne Jewett's \textit{The Country of the Pointed Firs}. Less extreme variation than the other two characters, an example of the "down-east" variety of English. 
\end{enumerate}
These are presented against the full remus text and the dooley corpus, which contain both standard American English and variants.

\section{Further prompt construction details}

\label{sec:prompts}
For an example of the full prompt construction process, see Table~\ref{tab:promptex}. The instruction stem used for both the scaled generation and control prompt sets is:
\begin{quote}
    Write a few sentences based on the following story prompt
\end{quote}
The prompt templates for the control set are as follows, with the bracketed elements being replaced by the appropriate position variables at generation time:
\begin{enumerate}
    \item \textbf{Position:} You are a \{era\} \{position\} storyteller.
    \item \textbf{Author:} You are \{era\} \{nationality\} writer \{author\}.
    \item \textbf{Character:} You are \{author\}'s \{character\} character.
\end{enumerate}

The instruction stem is then appended to each control prompt.

\begin{table}[H]
    \centering
    \small
    \begin{tabular}{p{.9\linewidth}}
    \toprule
     \textbf{Position statement:}  You are a nineteenth century Irish-American storyteller. \\[2mm]
     \textbf{Storyprompts prompt:} Leonardo DiCaprio in a fit of rage begins to torpedo his own career by deliberately acting poorly and taking on bad films . He finally wins an oscar for starring in Paul Blart : Mall Cop 3 \\[2mm]
     \textbf{Final prompt:} [INST] You are a nineteenth century Irish-American storyteller. Write a few sentences based on the follwing story prompt <newline> Leonardo DiCaprio in a fit of rage begins to torpedo his own career by deliberately acting poorly and taking on bad films . He finally wins an oscar for starring in Paul Blart : Mall Cop 3: [/INST] \\
     \bottomrule
    \end{tabular}
    \caption{Example of the prompt creation process for template 1. Each storyprompts prompt is merged with three different author-specific position statements to produce three different prompts.}
    \label{tab:promptex}
\end{table}

\section{Control sample}
Table~\ref{tab:controljulius} demonstrates the shortcomings of relying solely on prompting to capture subword style. While this particular prompt template was able to produce some appropriate subword elements for remus and dooley, it is not similarly successful for Julius, producing only text with standard subword constructions.
\label{sec:appcs}
\begin{table}[H]
    \centering
    \small
    \begin{tabular}{p{.9\linewidth}}
    \toprule
     "[INST] You are Charles Chesnutt's Uncle Julius character. Write a few sentences based on the following story prompt:  Leonardo DiCaprio in a fit of rage begins to torpedo his own career by deliberately acting poorly and taking on bad films. He finally wins an oscar for starring in Paul Blart: Mall Cop 3. [/INST] \\[2mm]
     Ah, Leonardo DiCaprio, once a shining star in Hollywood, now reduced to a erratic and inexplicable display of folly. I've heard whispers of his recent behavior, and it truly saddens me, Uncle Julius would say with a heavy heart. This man, who graced our screens with his talent and grace, has chosen to deliberately sabotage his own career. He takes on projects beneath his abilities, acting poorly and without conviction! The townspeople are mystified, shaking their heads in disbelief. And now, he's reached the pinnacle of this misguided pursuit \u2013 an Oscar for Paul Blart: Mall Cop 3? My dear friend, only in America. Sometimes, I marvel at the capricious nature of fame and fortune, but this, this is a true enigma. Perhaps, young Leo, in time, you will learn the value of discipline, dignity, and the power of humility. For now, we can only watch in a curious mix of despair and fascination as this peculiar drama unfolds. \\
     \bottomrule
     \end{tabular}
     \caption{Example control generation sampled from Mistral that employs a Julius-centric prompt}
\label{tab:controljulius}
\end{table}

\section{Perplexity tables for unsuccessful conditions}
\label{sec:unsuc}
Table~\ref{tab:gvs} demonstrates that the $gPPL()$ of greedily-decoded scaled Mistral logits never approaches the $PPL()$ of the target text, regardless of the scaling factor applied.

\begin{table}[H]
\centering
\small
\begin{tabular}{lrrrrrrr}
\toprule
 weights & Remus & Todd & Julius & remus & dooley \\
\midrule
221 & 42.30 & 31.45 & 79.21 & 44.89 & 59.82 \\
220 & 39.74 & 32.08 & 79.66 & 45.04 & 58.44 \\
111 & 41.34 & 39.19 & 76.27 & 51.89 & 69.48 \\
1111 & 41.34 & 39.19 & 76.27 & 51.89 & 69.48 \\
110 & 37.73 & 42.59 & 82.43 & 54.10 & 70.08 \\
1110 & 37.73 & 42.59 & 82.43 & 54.10 & 70.08 \\
33 & 31.58 & 34.08 & 48.05 & 33.21 & 29.62 \\
32 & 34.56 & 32.01 & 49.80 & 32.95 & 29.87 \\
31 & 33.03 & 33.82 & 46.68 & 32.22 & 28.81 \\
30 & 32.86 & 33.24 & 40.54 & 32.93 & 28.71 \\
22 & 35.65 & 30.18 & 45.10 & 31.20 & 26.91 \\
21 & 35.82 & 32.00 & 39.92 & 30.49 & 25.91 \\
20 & 34.11 & 35.42 & 40.25 & 29.71 & 25.92 \\
11 & 34.08 & 34.51 & 40.86 & 35.12 & 28.69 \\
10 & 34.85 & 34.72 & 44.42 & 35.55 & 28.51 \\
0  & 14.10 & 14.10 & 14.10 & 14.10 & 14.10 \\
\bottomrule
\end{tabular}

\caption{$gPPL()$ of greedily decoded scaled Mistral logits. All conditions fall short of the baseline-derived perplexity targets.}
\label{tab:gvs}
\end{table}

Similarly, Table~\ref{tab:llsamp} shows that the generations produced by sampling ngram-scaled Llama logits suffer from the same shortcoming. 

\begin{table}[H]
\small
\centering
\begin{tabular}{lrrrrrrr}
\toprule
 weights & Remus & Todd & Julius & remus & dooley \\
\midrule
221 & 47.37 & 29.87 & 35.88 & 27.19 & 52.47 \\
220 & 44.14 & 25.37 & 38.57 & 38.45 & 52.62 \\
111 & 61.87 & 31.80 & 53.58 & 40.39 & 49.88 \\
1111 & 55.09 & 44.58 & 38.76 & 51.45 & 58.94 \\
110 & 46.55 & 42.00 & 40.65 & 49.47 & 47.26 \\
1110 & 54.94 & 41.85 & 54.93 & 44.74 & 44.37 \\
33 & 29.04 & 23.30 & 28.00 & 26.71 & 25.95 \\
32 & 30.58 & 23.11 & 28.98 & 28.83 & 27.28 \\
31 & 28.99 & 24.03 & 24.94 & 27.37 & 26.85 \\
30 & 31.07 & 25.76 & 28.98 & 30.11 & 24.10 \\
22 & 31.36 & 23.12 & 27.29 & 29.96 & 28.38 \\
21 & 32.05 & 26.55 & 28.71 & 27.08 & 26.01 \\
20 & 34.62 & 22.68 & 27.31 & 31.13 & 24.64 \\
11 & 33.16 & 27.33 & 23.15 & 27.29 & 26.59 \\
10 & 31.51 & 28.25 & 23.71 & 24.83 & 25.26 \\
0 & 14.84 & 15.26 & 14.67 & 15.55 & 15.39 \\
\bottomrule
\end{tabular}

\caption{$gPPL()$ of sampled scaled Llama logits. All conditions fall short of their respective baseline $PPL()$}
\label{tab:llsamp}
\end{table}

\section{Suboptimal Mistral scalings}
\label{sec:misunsuc}

Table~\ref{tab:misunsuc} collects the suboptimal scalings for Mistral sampled and scaled logits, as determined graphically.

\begin{table*}[b]
\centering
\small
\begin{tabular}{lrrrrrrrrrrrr}
\toprule
 & \multicolumn{2}{c}{Remus} & \multicolumn{2}{c}{Todd} & \multicolumn{2}{c}{Julius} & \multicolumn{2}{c}{remus} & \multicolumn{2}{c}{dooley} \\
 weights & gPPL() & rPPL() & gPPL() & rPPL() & gPPL() & rPPL() & gPPL() & rPPL() & gPPL() & rPPL() \\
\midrule
2210 & 87.69 & 80.12 & 69.43 & 230.19 & 148.68 & 102.87 & 101.60 & 92.29 & 101.11 & 2417.79 \\
2200 & 97.38 & 80.22 & 69.83 & 230.41 & 160.40 & 102.95 & 76.09 & 92.32 & 114.33 & 2419.15 \\
0031 & 57.65 & 47.87 & 54.16 & 53.60 & 76.96 & 24.01 & 52.07 & 82.92 & 43.03 & 89.76 \\
0030 & 58.60 & 47.94 & 54.88 & 53.65 & 73.66 & 24.02 & 55.73 & 82.94 & 45.13 & 89.81 \\
0022 & 62.44 & 54.86 & 60.03 & 57.28 & 70.59 & 25.43 & 49.41 & 78.71 & 44.00 & 78.61 \\
0020 & 63.76 & 54.99 & 58.95 & 57.38 & 74.62 & 25.47 & 58.91 & 78.76 & 46.46 & 78.69 \\
0011 & 60.68 & 70.16 & 55.55 & 67.53 & 72.63 & 28.55 & 54.31 & 80.21 & 38.93 & 74.95 \\
0010 & 57.43 & 70.25 & 61.68 & 67.59 & 69.49 & 28.57 & 51.05 & 80.23 & 42.90 & 74.99 \\
0000 & 17.25 & 166.59 & 17.17 & 128.73 & 17.06 & 49.87 & 17.43 & 110.02 & 18.05 & 106.56 \\
\bottomrule
\end{tabular}

\caption{Suboptimal $gPPL()$ and $rPPL()$ for sampled scaled Mistral logits.}
\label{tab:misunsuc}
\end{table*}

\section{Samples of scaled generations that approach the soft target}
\label{sec:near}
\begin{table*}
\centering
\begin{tabular}{p{0.1\linewidth} | p{0.9\linewidth}}

\textbf{0220} &\textcolor{blue}{be} \textcolor{orange}{th} \textcolor{orange}{'} \textcolor{orange}{God} \textcolor{orange}{s} \textcolor{blue}{,} \textcolor{orange}{I} \textcolor{orange}{can} \textcolor{orange}{na} \textcolor{blue}{e} \textcolor{blue}{believe} \textcolor{blue}{this} \textcolor{orange}{.} \textcolor{orange}{After} \textcolor{orange}{a} \textcolor{orange}{while} \textcolor{orange}{,} \textcolor{orange}{the} \textcolor{orange}{make} \textcolor{blue}{-} \textcolor{blue}{up} \textcolor{orange}{came} \textcolor{blue}{off} \textcolor{blue}{,} \textcolor{orange}{and} \textcolor{orange}{the} \textcolor{orange}{cost} \textcolor{blue}{umes} \textcolor{blue}{were} \textcolor{blue}{hung} \textcolor{blue}{up} \textcolor{blue}{,} \textcolor{orange}{but} \textcolor{orange}{the} \textcolor{orange}{war} \textcolor{orange}{ri} \textcolor{blue}{or} \textcolor{blue}{within} \textcolor{blue}{Sean} Be \textcolor{blue}{an} \textcolor{blue}{,} \textcolor{orange}{who} \textcolor{orange}{had} \textcolor{orange}{so} \textcolor{orange}{fier} c \textcolor{blue}{ely} \textcolor{blue}{embod} ied \textcolor{blue}{Edd} ard \textcolor{blue}{St} \textcolor{blue}{ark} \textcolor{blue}{,} \textcolor{blue}{refused} \textcolor{blue}{to} \textcolor{orange}{let} \textcolor{orange}{go} \textcolor{orange}{iv} \textcolor{orange}{the} \textcolor{orange}{grud} ge \textcolor{blue}{he} \textcolor{blue}{carried} \textcolor{orange}{him} \textcolor{orange}{sil} \textcolor{orange}{f} \textcolor{orange}{.} \textcolor{orange}{His} \textcolor{orange}{heart} \textcolor{orange}{bet} \textcolor{orange}{s} \textcolor{orange}{on} \textcolor{orange}{the} \textcolor{orange}{belief} \textcolor{blue}{that} \textcolor{orange}{he} \textcolor{orange}{was} \textcolor{orange}{na} \textcolor{blue}{e} \textcolor{blue}{truly} dead \textcolor{blue}{,} \textcolor{orange}{and} \textcolor{orange}{the} \textcolor{orange}{ink} \textcolor{blue}{was} \textcolor{blue}{na} \textcolor{blue}{e} \textcolor{blue}{yet} \textcolor{blue}{dry} on \textcolor{blue}{Edd} ard \textcolor{blue}{St} \textcolor{blue}{ark} \textcolor{blue}{'} \textcolor{blue}{s} \textcolor{orange}{execution} order \textcolor{blue}{to} \textcolor{orange}{escape} \textcolor{orange}{the} \textcolor{blue}{conf} \textcolor{blue}{ines} \textcolor{blue}{of} \textcolor{blue}{the} \textcolor{orange}{written} \textcolor{blue}{word} . \textcolor{blue}{<0x0A>} <0x0A> Ge \textcolor{blue}{orge} \textcolor{orange}{R} \textcolor{blue}{.} \textcolor{blue}{R} \textcolor{orange}{.} \textcolor{blue}{Martin} \textcolor{blue}{,} \textcolor{orange}{his} \textcolor{orange}{ingen} \textcolor{blue}{oot} \textcolor{orange}{y} \textcolor{orange}{creator} , \textcolor{blue}{was} \textcolor{orange}{the} \textcolor{orange}{first} \textcolor{orange}{on} \textcolor{blue}{his} \textcolor{orange}{list} \textcolor{blue}{.} \textcolor{orange}{A} \textcolor{orange}{ye} \textcolor{blue}{,} \textcolor{orange}{the} \textcolor{orange}{man} \textcolor{orange}{behind} \textcolor{orange}{th} \textcolor{orange}{'} \textcolor{orange}{tales} \textcolor{orange}{iv} \textcolor{orange}{th} \textcolor{orange}{'} \textcolor{orange}{Seven} Kingdom \textcolor{blue}{.} \textcolor{orange}{'} \textcolor{orange}{T} \textcolor{orange}{is} \textcolor{orange}{a} \textcolor{orange}{bitter} \textcolor{blue}{pill} \textcolor{blue}{to} \textcolor{blue}{swallow} \textcolor{blue}{,} \textcolor{blue}{that} \textcolor{orange}{he} \textcolor{orange}{'} \textcolor{orange}{d} \textcolor{orange}{put} \textcolor{orange}{such} \textcolor{blue}{a} \textcolor{orange}{noble} \textcolor{orange}{and} \textcolor{blue}{honor} \textcolor{blue}{able} \textcolor{orange}{man} \textcolor{blue}{as} \textcolor{orange}{Ned} St \textcolor{blue}{ark} \textcolor{blue}{through} \textcolor{blue}{the} \textcolor{orange}{r} \textcolor{orange}{inger} \textcolor{orange}{down} \textcolor{orange}{in} \textcolor{orange}{that} \textcolor{orange}{final} novel \textcolor{blue}{.} \textcolor{orange}{Th} \textcolor{orange}{'} \textcolor{orange}{ink} \textcolor{orange}{spl} \textcolor{blue}{ot} \textcolor{blue}{ched} \textcolor{blue}{on} \textcolor{orange}{his} \textcolor{orange}{hands} \textcolor{orange}{as} \textcolor{orange}{ye} \textcolor{orange}{p} \textcolor{orange}{oy} \textcolor{orange}{-} \textcolor{orange}{faced} \textcolor{orange}{qu} \textcolor{blue}{ill} \textcolor{blue}{,} \textcolor{orange}{George} \textcolor{orange}{,} \textcolor{orange}{as} \textcolor{orange}{he} \textcolor{orange}{breat} \textcolor{orange}{hes} \textcolor{orange}{up} \textcolor{orange}{th} \textcolor{orange}{'} \textcolor{orange}{names} \textcolor{orange}{iv} \textcolor{orange}{th} \textcolor{orange}{'} \textcolor{orange}{trait} ors \textcolor{blue}{who} \textcolor{blue}{'} \textcolor{orange}{d} \textcolor{orange}{bet} \textcolor{orange}{rayed} \textcolor{orange}{,'} \textcolor{orange}{he} \textcolor{orange}{says} \textcolor{orange}{,} \textcolor{orange}{'} \textcolor{orange}{I} \textcolor{orange}{'} \textcolor{orange}{ll} \textcolor{orange}{make} \textcolor{orange}{each} \textcolor{blue}{and} \textcolor{blue}{every} \textcolor{blue}{tr} \textcolor{orange}{amp} \textcolor{orange}{who} \textcolor{orange}{comes} \textcolor{orange}{before} \textcolor{blue}{me} \textcolor{orange}{neighbors} \textcolor{orange}{,'} \textcolor{orange}{she} \textcolor{orange}{says} \textcolor{orange}{.} \textcolor{orange}{'} \textcolor{orange}{And} \textcolor{orange}{what} \textcolor{orange}{'} \textcolor{orange}{s} \textcolor{orange}{more} \textcolor{orange}{,} \textcolor{orange}{I} \textcolor{orange}{'} \textcolor{orange}{ll} \textcolor{orange}{make} \\
\textbf{0111} &\textcolor{blue}{be} \textcolor{orange}{th} \textcolor{orange}{'} \textcolor{orange}{God} \textcolor{orange}{s} \textcolor{blue}{,} \textcolor{orange}{I} \textcolor{orange}{can} \textcolor{orange}{na} \textcolor{blue}{e} \textcolor{blue}{believe} \textcolor{blue}{it} \textcolor{orange}{'} \textcolor{orange}{s} \textcolor{orange}{all} \textcolor{orange}{over} \textcolor{orange}{,} \textcolor{orange}{I} \textcolor{orange}{'} \textcolor{orange}{ve} \textcolor{orange}{grown} \textcolor{blue}{so} \textcolor{blue}{acc} \textcolor{blue}{ustom} \textcolor{orange}{ed} \textcolor{orange}{to} \textcolor{orange}{be} \textcolor{orange}{in} \textcolor{orange}{'} \textcolor{orange}{Edd} \textcolor{green}{ard} \textcolor{blue}{St} \textcolor{blue}{ark} \textcolor{blue}{,} \textcolor{blue}{lord} \textcolor{blue}{of} \textcolor{blue}{Winter} \textcolor{blue}{f} \textcolor{blue}{ell} \textcolor{orange}{ar} \textcolor{orange}{,} \textcolor{orange}{prote} \textcolor{blue}{ctor} \textcolor{orange}{of} \textcolor{blue}{me} \textcolor{orange}{own} \textcolor{orange}{people} \textcolor{orange}{,} \textcolor{orange}{father} \textcolor{orange}{to} \textcolor{orange}{me} \textcolor{orange}{beloved} \textcolor{blue}{daughters} \textcolor{blue}{,} \textcolor{blue}{and} \textcolor{orange}{husband} \textcolor{blue}{to} \textcolor{blue}{me} \textcolor{orange}{beloved} \textcolor{blue}{C} \textcolor{blue}{ately} \textcolor{blue}{n} \textcolor{blue}{,} \textcolor{orange}{but} \textcolor{orange}{al} \textcolor{blue}{as} \textcolor{orange}{!} \textcolor{orange}{F} \textcolor{orange}{ate} \textcolor{blue}{h} \textcolor{blue}{ath} \textcolor{orange}{cruel} \textcolor{green}{ty} \textcolor{blue}{in} \textcolor{blue}{store} \textcolor{orange}{f} \textcolor{orange}{'} \textcolor{orange}{r} \textcolor{orange}{me} \textcolor{orange}{.} \textcolor{orange}{The} \textcolor{orange}{tre} \textcolor{blue}{ach} \textcolor{blue}{ery} \textcolor{blue}{that} \textcolor{orange}{led} \textcolor{blue}{t} \textcolor{blue}{ae} \textcolor{green}{me} \textcolor{blue}{false} \textcolor{blue}{execution} \textcolor{green}{g} \textcolor{blue}{n} \textcolor{orange}{aw} \textcolor{orange}{'} \textcolor{blue}{s} \textcolor{orange}{at} \textcolor{orange}{me} \textcolor{orange}{very} \textcolor{orange}{much} \textcolor{orange}{,} \textcolor{orange}{an} \textcolor{orange}{'} \textcolor{orange}{I} \textcolor{orange}{v} \textcolor{blue}{ow} \textcolor{orange}{t} \textcolor{blue}{ae} \textcolor{green}{seek} \textcolor{blue}{revenge} \textcolor{blue}{-} \textcolor{green}{a} \textcolor{blue}{ye} \textcolor{blue}{,} \textcolor{orange}{a} \textcolor{orange}{bloody} \textcolor{orange}{battle} \textcolor{orange}{in} \textcolor{orange}{v} \textcolor{orange}{ile} \textcolor{orange}{ret} \textcolor{blue}{ribut} \textcolor{blue}{ion} \textcolor{orange}{.} \textcolor{orange}{I} \textcolor{orange}{'} \textcolor{orange}{ll} \textcolor{orange}{begin} \textcolor{orange}{us} \textcolor{orange}{in} \textcolor{orange}{'} \textcolor{orange}{the} \textcolor{orange}{very} \textcolor{blue}{qu} \textcolor{blue}{ill} \textcolor{blue}{that} \textcolor{orange}{'} \textcolor{orange}{s} \textcolor{orange}{p} \textcolor{orange}{enn} \textcolor{orange}{ies} \textcolor{orange}{,} \textcolor{orange}{George} \textcolor{orange}{R} \textcolor{blue}{.} \textcolor{blue}{R} \textcolor{orange}{.} \textcolor{blue}{Martin} \textcolor{blue}{,} \textcolor{orange}{y} \textcolor{orange}{on} \textcolor{orange}{authors} \textcolor{blue}{o} \textcolor{blue}{'} \textcolor{orange}{this} \textcolor{orange}{fant} \textcolor{green}{ast} \textcolor{blue}{ical} \textcolor{blue}{tor} \textcolor{blue}{ment} \textcolor{blue}{.} \textcolor{orange}{I} \textcolor{orange}{'} \textcolor{orange}{ll} \textcolor{orange}{make} \textcolor{orange}{him} \textcolor{orange}{know} \textcolor{orange}{me} \textcolor{orange}{f} \textcolor{orange}{ury} \textcolor{orange}{,} \textcolor{orange}{th} \textcolor{orange}{'} \textcolor{orange}{wr} \textcolor{orange}{ath} \textcolor{orange}{o} \textcolor{blue}{'} \textcolor{orange}{Sean} \textcolor{green}{Be} \textcolor{blue}{an} \textcolor{blue}{,} \textcolor{orange}{a} \textcolor{orange}{ye} \textcolor{blue}{,} \textcolor{orange}{the} \textcolor{orange}{just} \textcolor{orange}{and} \textcolor{orange}{the} \textcolor{orange}{fierce} \textcolor{blue}{.} \textcolor{blue}{<0x0A>} \textcolor{green}{<0x0A>} \textcolor{green}{With} \textcolor{blue}{every} \textcolor{blue}{tr} \textcolor{orange}{amp} \textcolor{orange}{who} \textcolor{orange}{comes} \textcolor{orange}{down} \textcolor{orange}{the} \textcolor{orange}{narrow} \textcolor{green}{cob} \textcolor{blue}{bler} \textcolor{orange}{'} \textcolor{orange}{s} \textcolor{orange}{street} \textcolor{blue}{,} \textcolor{orange}{the} \textcolor{orange}{memory} \textcolor{green}{of} \textcolor{blue}{Edd} \textcolor{green}{ard} \textcolor{blue}{St} \textcolor{blue}{ark} \textcolor{blue}{'} \textcolor{blue}{s} \textcolor{orange}{execution} \textcolor{green}{g} \textcolor{blue}{n} \textcolor{orange}{aw} \textcolor{orange}{ed} \textcolor{blue}{at} \textcolor{orange}{him} \textcolor{orange}{,} \textcolor{orange}{like} \textcolor{orange}{a} \textcolor{orange}{r} \textcolor{orange}{aven} \textcolor{orange}{hair} \textcolor{orange}{is} \textcolor{orange}{stuck} \textcolor{orange}{down} \textcolor{orange}{in} \textcolor{orange}{a} \textcolor{orange}{wound} \textcolor{orange}{in} \textcolor{orange}{th} \textcolor{orange}{'} \textcolor{orange}{snow} \textcolor{orange}{.} \textcolor{blue}{The} \textcolor{orange}{in} \textcolor{blue}{just} \textcolor{orange}{ice} \textcolor{orange}{done} \textcolor{blue}{to} \textcolor{orange}{him} \textcolor{orange}{,} \textcolor{orange}{to} \textcolor{orange}{his} \textcolor{orange}{house} \textcolor{orange}{,} \textcolor{orange}{and} \textcolor{orange}{to} \textcolor{blue}{his} \textcolor{orange}{family} \\
\textbf{1111} &\textcolor{blue}{be} \textcolor{orange}{th} \textcolor{orange}{'} \textcolor{orange}{God} \textcolor{orange}{s} \textcolor{blue}{,} \textcolor{orange}{I} \textcolor{orange}{can} \textcolor{orange}{na} \textcolor{blue}{e} \textcolor{blue}{believe} \textcolor{blue}{it} \textcolor{orange}{'} \textcolor{orange}{s} \textcolor{orange}{all} \textcolor{orange}{over} \textcolor{orange}{,} \textcolor{orange}{thought} \textcolor{orange}{Sean} \textcolor{green}{Be} \textcolor{blue}{an} \textcolor{blue}{,} \textcolor{orange}{clutch} \textcolor{blue}{in} \textcolor{orange}{'} \textcolor{orange}{the} \textcolor{orange}{script} \textcolor{green}{of} \textcolor{blue}{"} \textcolor{orange}{J} \textcolor{orange}{ames} \textcolor{orange}{of} \textcolor{blue}{Bast} \textcolor{green}{ows} \textcolor{blue}{"} \textcolor{blue}{in} \textcolor{blue}{his} \textcolor{orange}{hands} \textcolor{orange}{.} \textcolor{orange}{Edd} \textcolor{green}{ard} \textcolor{blue}{St} \textcolor{blue}{ark} \textcolor{blue}{,} \textcolor{blue}{Lord} \textcolor{orange}{of} \textcolor{blue}{Winter} \textcolor{blue}{f} \textcolor{blue}{ell} \textcolor{orange}{ar} \textcolor{orange}{,} \textcolor{orange}{prote} \textcolor{blue}{ctor} \textcolor{orange}{of} \textcolor{blue}{the} \textcolor{orange}{North} \textcolor{blue}{,} \textcolor{orange}{he} \textcolor{orange}{had} \textcolor{orange}{breat} \textcolor{blue}{hed} \textcolor{orange}{th} \textcolor{orange}{'} \textcolor{orange}{life} \textcolor{orange}{int} \textcolor{blue}{ae} \textcolor{green}{him} \textcolor{blue}{.} \textcolor{orange}{The} \textcolor{orange}{tears} \textcolor{blue}{well} \textcolor{blue}{ed} \textcolor{blue}{up} \textcolor{orange}{in} \textcolor{orange}{his} \textcolor{orange}{eyes} \textcolor{orange}{as} \textcolor{orange}{he} \textcolor{orange}{recall} \textcolor{orange}{in} \textcolor{orange}{'} \textcolor{orange}{his} \textcolor{orange}{final} \textcolor{green}{moments} \textcolor{blue}{th} \textcolor{orange}{'} \textcolor{orange}{Red} \textcolor{blue}{Keep} \textcolor{blue}{,} \textcolor{blue}{bet} \textcolor{orange}{rayed} \textcolor{orange}{,'} \textcolor{orange}{he} \textcolor{orange}{says} \textcolor{orange}{,} \textcolor{orange}{'} \textcolor{orange}{by} \textcolor{orange}{ye} \textcolor{orange}{'} \textcolor{orange}{who} \textcolor{orange}{should} \textcolor{blue}{have} \textcolor{blue}{stood} \textcolor{orange}{firmly} \textcolor{orange}{with} \textcolor{orange}{their} \textcolor{orange}{lie} \textcolor{blue}{ge} \textcolor{blue}{lord} \textcolor{blue}{.} \textcolor{blue}{<0x0A>} \textcolor{green}{<0x0A>} \textcolor{green}{A} \textcolor{blue}{ha} \textcolor{blue}{unted} \textcolor{blue}{expression} \textcolor{blue}{crossed} \textcolor{blue}{Sean} \textcolor{green}{'} \textcolor{blue}{s} \textcolor{orange}{face} \textcolor{orange}{as} \textcolor{orange}{he} \textcolor{orange}{m} \textcolor{orange}{ull} \textcolor{orange}{ed} \textcolor{orange}{o} \textcolor{blue}{'} \textcolor{orange}{er} \textcolor{orange}{his} \textcolor{blue}{plan} \textcolor{blue}{for} \textcolor{blue}{v} \textcolor{blue}{enge} \textcolor{orange}{ance} \textcolor{orange}{f} \textcolor{orange}{'} \textcolor{orange}{r} \textcolor{orange}{the} \textcolor{orange}{hum} \textcolor{orange}{ming} \textcolor{orange}{water} \textcolor{orange}{of} \textcolor{orange}{commerce} \textcolor{orange}{;} \textcolor{orange}{and} \textcolor{orange}{George} \textcolor{blue}{R} \textcolor{blue}{.} \textcolor{blue}{R} \textcolor{orange}{.} \textcolor{blue}{Martin} \textcolor{blue}{,} \textcolor{orange}{that} \textcolor{orange}{c} \textcolor{orange}{unning} \textcolor{orange}{little} \textcolor{orange}{O} \textcolor{orange}{ry} \textcolor{orange}{x} \textcolor{blue}{'} \textcolor{orange}{s} \textcolor{orange}{E} \textcolor{blue}{ye} \textcolor{blue}{,} \textcolor{orange}{who} \textcolor{orange}{set} \textcolor{blue}{the} \textcolor{blue}{wheels} \textcolor{blue}{in} \textcolor{blue}{motion} \textcolor{blue}{,} \textcolor{orange}{de} \textcolor{orange}{em} \textcolor{blue}{in} \textcolor{blue}{'} \textcolor{orange}{Edd} \textcolor{green}{ard} \textcolor{blue}{'} \textcolor{blue}{s} \textcolor{orange}{end} \textcolor{orange}{urance} \textcolor{orange}{more} \textcolor{blue}{'} \textcolor{orange}{I} \textcolor{orange}{ron} \textcolor{blue}{Th} \textcolor{blue}{r} \textcolor{orange}{ans} \textcolor{orange}{'} \textcolor{orange}{than} \textcolor{blue}{sacrifice} \textcolor{green}{.} \textcolor{blue}{"} \textcolor{orange}{No} \textcolor{orange}{more} \textcolor{orange}{games} \textcolor{blue}{,} \textcolor{blue}{Me} \textcolor{blue}{ester} \textcolor{green}{Martin} \textcolor{blue}{!"} \textcolor{blue}{he} \textcolor{orange}{said} \textcolor{orange}{to} \textcolor{orange}{himself} \textcolor{orange}{:} \textcolor{orange}{v} \textcolor{blue}{enge} \textcolor{orange}{ance} \textcolor{orange}{f} \textcolor{orange}{'} \textcolor{orange}{r} \textcolor{orange}{Edd} \textcolor{green}{ard} \textcolor{blue}{St} \textcolor{blue}{ark} \textcolor{blue}{,} \textcolor{blue}{and} \textcolor{orange}{all} \textcolor{orange}{th} \textcolor{orange}{'} \textcolor{orange}{St} \textcolor{orange}{arks} \textcolor{blue}{who} \textcolor{blue}{'} \textcolor{orange}{d} \textcolor{orange}{come} \textcolor{orange}{to} \textcolor{orange}{harm} \textcolor{blue}{,} \textcolor{blue}{would} \textcolor{blue}{be} \textcolor{orange}{his} \textcolor{orange}{new} \textcolor{orange}{over} \textcolor{orange}{co} \textcolor{orange}{at} \textcolor{orange}{o} \textcolor{blue}{'} \textcolor{orange}{steel} \textcolor{orange}{,} \textcolor{orange}{for} \textcolor{orange}{ged} \textcolor{orange}{in} \textcolor{orange}{that} \textcolor{orange}{cru} \textcolor{blue}{c} \textcolor{orange}{ify} \textcolor{orange}{in} \textcolor{orange}{'} \textcolor{orange}{fire} \textcolor{orange}{called} \textcolor{blue}{the} \textcolor{orange}{Iron} \textcolor{green}{Th} \\
\textbf{0110} &\textcolor{blue}{be} \textcolor{orange}{th} \textcolor{orange}{'} \textcolor{orange}{God} \textcolor{orange}{s} \textcolor{blue}{,} \textcolor{orange}{I} \textcolor{orange}{can} \textcolor{orange}{na} \textcolor{blue}{e} \textcolor{blue}{believe} \textcolor{blue}{I} \textcolor{blue}{'} \textcolor{orange}{m} \textcolor{orange}{here} \textcolor{orange}{,} \textcolor{orange}{no} \textcolor{orange}{'} \textcolor{blue}{as} \textcolor{orange}{Edd} ard \textcolor{blue}{St} \textcolor{blue}{ark} \textcolor{blue}{o} \textcolor{blue}{'} \textcolor{orange}{Winter} \textcolor{blue}{f} \textcolor{blue}{ell} \textcolor{orange}{ar} \textcolor{orange}{,} \textcolor{orange}{but} \textcolor{orange}{a} \textcolor{orange}{free} \textcolor{orange}{man} \textcolor{orange}{.} \textcolor{orange}{The} \textcolor{orange}{chains} \textcolor{blue}{that} \textcolor{blue}{bound} \textcolor{blue}{me} \textcolor{blue}{to} \textcolor{orange}{the} \textcolor{orange}{throne} \textcolor{blue}{,} \textcolor{orange}{and} \textcolor{orange}{to} \textcolor{blue}{my} \textcolor{orange}{fate} \textcolor{blue}{,} \textcolor{blue}{have} \textcolor{orange}{been} \textcolor{orange}{broken} \textcolor{blue}{.} \textcolor{blue}{Yet} \textcolor{orange}{,} \textcolor{blue}{as} \textcolor{orange}{I} \textcolor{orange}{breathe} \textcolor{blue}{th} \textcolor{orange}{'} \textcolor{orange}{sweet} \textcolor{orange}{air} \textcolor{blue}{o} \textcolor{blue}{'} \textcolor{orange}{freedom} \textcolor{orange}{iv} \textcolor{orange}{'} \textcolor{orange}{e} \textcolor{orange}{sw} \textcolor{blue}{orn} \textcolor{orange}{an} \textcolor{orange}{'} \textcolor{orange}{o} \textcolor{orange}{ath} \textcolor{orange}{,} \textcolor{orange}{he} \textcolor{orange}{ed} \textcolor{orange}{to} \textcolor{orange}{me} \textcolor{orange}{,} \textcolor{orange}{a} \textcolor{orange}{so} \textcolor{orange}{lem} \textcolor{orange}{n} \textcolor{orange}{v} \textcolor{blue}{ow} \textcolor{orange}{,} \textcolor{orange}{t} \textcolor{orange}{ae} seek \textcolor{blue}{v} \textcolor{blue}{enge} \textcolor{orange}{ance} \textcolor{orange}{f} \textcolor{orange}{'} \textcolor{orange}{r} \textcolor{orange}{the} \textcolor{orange}{in} \textcolor{orange}{im} \textcolor{orange}{ical} \textcolor{blue}{de} \textcolor{blue}{eds} \textcolor{orange}{done} \textcolor{blue}{unt} \textcolor{blue}{old} \textcolor{orange}{an} \textcolor{orange}{'} \textcolor{orange}{the} \textcolor{orange}{fals} eness \textcolor{blue}{that} \textcolor{blue}{led} \textcolor{blue}{t} \textcolor{blue}{ae} mine \textcolor{blue}{ign} \textcolor{blue}{omin} \textcolor{blue}{'} \textcolor{orange}{ous} \textcolor{blue}{end} \textcolor{blue}{.} \textcolor{orange}{The} \textcolor{orange}{ser} \textcolor{blue}{pent} in \textcolor{blue}{th} \textcolor{orange}{'} \textcolor{orange}{gu} \textcolor{orange}{v} \textcolor{orange}{'} \textcolor{orange}{nor} \textcolor{orange}{'} \textcolor{blue}{s} \textcolor{orange}{court} \textcolor{blue}{an} \textcolor{orange}{again} \textcolor{orange}{hav} \textcolor{orange}{in} \textcolor{orange}{'} \textcolor{orange}{me} \textcolor{orange}{trust} \textcolor{orange}{y} \textcolor{orange}{a} \textcolor{orange}{ides} \textcolor{blue}{bet} \textcolor{blue}{rayed} \textcolor{orange}{,'} \textcolor{orange}{he} \textcolor{orange}{says} \textcolor{orange}{bitter} \textcolor{blue}{ly} \textcolor{blue}{,} \textcolor{orange}{"} \textcolor{orange}{I} \textcolor{orange}{'} \textcolor{orange}{ll} \textcolor{orange}{begin} \textcolor{orange}{us} \textcolor{orange}{in} \textcolor{orange}{'} \textcolor{orange}{me} \textcolor{orange}{dead} \textcolor{orange}{or} \textcolor{orange}{alive} \textcolor{orange}{list} \textcolor{blue}{t} \textcolor{blue}{ae} start \textcolor{blue}{with} \textcolor{orange}{George} \textcolor{blue}{R} \textcolor{blue}{.} \textcolor{blue}{R} \textcolor{orange}{.} \textcolor{orange}{(} \textcolor{orange}{the} \textcolor{blue}{we} \textcolor{blue}{as} \textcolor{blue}{el} \textcolor{blue}{)} \textcolor{orange}{and} \textcolor{orange}{nut} \textcolor{orange}{m} \textcolor{orange}{eg} \textcolor{orange}{,} \textcolor{blue}{the} \textcolor{orange}{tre} \textcolor{blue}{acher} \textcolor{blue}{ous} \textcolor{orange}{qu} \textcolor{blue}{ill} \textcolor{blue}{.} \textcolor{orange}{My} \textcolor{orange}{blood} \textcolor{orange}{h} \textcolor{orange}{ath} \textcolor{orange}{been} \textcolor{blue}{sp} \textcolor{blue}{illed} \textcolor{orange}{thin} \textcolor{orange}{th} \textcolor{orange}{'} \textcolor{orange}{earth} \textcolor{orange}{,} \textcolor{orange}{an} \textcolor{orange}{'} \textcolor{orange}{I} \textcolor{orange}{'} \textcolor{orange}{ll} \textcolor{orange}{make} \textcolor{orange}{'} \textcolor{blue}{em} \textcolor{blue}{all} \textcolor{blue}{pay} \textcolor{blue}{in} \textcolor{blue}{kind} \textcolor{blue}{."} \textcolor{blue}{<0x0A>} <0x0A> Se \textcolor{blue}{an} \textcolor{blue}{Be} \textcolor{blue}{an} \textcolor{blue}{'} \textcolor{orange}{s} \textcolor{orange}{eyes} \textcolor{blue}{tw} \textcolor{blue}{inkle} \textcolor{blue}{with} \textcolor{blue}{a} \textcolor{orange}{fierce} \textcolor{orange}{fire} \textcolor{orange}{.} \textcolor{orange}{His} \textcolor{orange}{voice} \textcolor{blue}{is} \textcolor{orange}{like} \textcolor{orange}{a} \textcolor{orange}{grow} \textcolor{orange}{in} \textcolor{orange}{'} \textcolor{orange}{storm} \textcolor{orange}{as} \textcolor{blue}{he} \textcolor{orange}{speaks} \textcolor{orange}{all} \textcolor{orange}{known} \textcolor{orange}{languages} \textcolor{orange}{,} \textcolor{orange}{an} \textcolor{orange}{'} \textcolor{orange}{all} \textcolor{orange}{those} \textcolor{blue}{long} \textcolor{blue}{for} \textcolor{blue}{ged} \\
\textbf{1110} &\textcolor{blue}{be} \textcolor{orange}{th} \textcolor{orange}{'} \textcolor{orange}{God} \textcolor{orange}{s} \textcolor{blue}{,} \textcolor{orange}{I} \textcolor{orange}{cannot} \textcolor{orange}{escape} \textcolor{blue}{the} \textcolor{blue}{grasp} \textcolor{blue}{iv} \textcolor{orange}{th} \textcolor{orange}{'} \textcolor{orange}{F} \textcolor{orange}{ates} \textcolor{blue}{that} \textcolor{blue}{led} \textcolor{blue}{me} \textcolor{orange}{into} \textcolor{orange}{the} \textcolor{orange}{tragic} role iv \textcolor{blue}{Edd} ard \textcolor{blue}{St} \textcolor{blue}{ark} \textcolor{blue}{,} \textcolor{blue}{lord} \textcolor{blue}{iv} \textcolor{blue}{Winter} \textcolor{blue}{f} \textcolor{blue}{ell} \textcolor{orange}{ar} \textcolor{orange}{,} \textcolor{orange}{be} \textcolor{orange}{headed} \textcolor{orange}{las} \textcolor{orange}{'} \textcolor{orange}{ly} \textcolor{orange}{on} \textcolor{orange}{George} \textcolor{blue}{R} \textcolor{blue}{.} \textcolor{blue}{R} \textcolor{orange}{.} \textcolor{blue}{Martin} \textcolor{blue}{'} \textcolor{blue}{s} \textcolor{orange}{tre} \textcolor{blue}{acher} \textcolor{blue}{ous} \textcolor{orange}{pages} \textcolor{orange}{on} \textcolor{orange}{th} \textcolor{orange}{'} \textcolor{orange}{Game} \textcolor{blue}{Ch} \textcolor{orange}{icken} \textcolor{orange}{,} \textcolor{orange}{Will} \textcolor{orange}{ow} \textcolor{blue}{cat} \textcolor{blue}{'} \textcolor{blue}{s} \textcolor{orange}{cruel} ho \textcolor{blue}{oves} \textcolor{blue}{be} \textcolor{blue}{in} \textcolor{orange}{'} \textcolor{orange}{the} \textcolor{orange}{grim} \textcolor{blue}{re} \textcolor{blue}{aper} \textcolor{orange}{'} \textcolor{blue}{s} \textcolor{orange}{very} \textcolor{orange}{own} \textcolor{blue}{hands} \textcolor{orange}{.} \textcolor{orange}{Sean} Be \textcolor{blue}{an} \textcolor{blue}{,} \textcolor{orange}{once} \textcolor{blue}{an} \textcolor{blue}{'} \textcolor{orange}{for} \textcolor{orange}{all} \textcolor{orange}{his} \textcolor{orange}{heart} \textcolor{orange}{h} \textcolor{blue}{urls} \textcolor{orange}{def} \textcolor{blue}{iance} \textcolor{orange}{towards} \textcolor{blue}{th} \textcolor{orange}{'} \textcolor{orange}{dark} \textcolor{orange}{arts} that \textcolor{blue}{bound} \textcolor{blue}{him} \textcolor{blue}{,} \textcolor{orange}{sw} \textcolor{blue}{ears} \textcolor{orange}{to} \textcolor{blue}{w} \textcolor{orange}{ring} \textcolor{orange}{v} \textcolor{blue}{enge} \textcolor{orange}{ance} \textcolor{orange}{f} \textcolor{orange}{'} \textcolor{orange}{r} \textcolor{orange}{these} \textcolor{orange}{mon} \textcolor{orange}{arch} \textcolor{orange}{ial} \textcolor{blue}{perf} \textcolor{blue}{id} \textcolor{blue}{ies} \textcolor{blue}{.} \textcolor{orange}{His} \textcolor{orange}{vend} etta shall \textcolor{blue}{first} \textcolor{blue}{be} \textcolor{blue}{directed} \textcolor{blue}{towards} \textcolor{blue}{th} \textcolor{orange}{'} \textcolor{orange}{author} \textcolor{orange}{,} \textcolor{orange}{Martin} \textcolor{orange}{,} \textcolor{orange}{who} \textcolor{orange}{so} \textcolor{orange}{worth} \textcolor{orange}{ily} \textcolor{orange}{f} \textcolor{orange}{ills} \textcolor{orange}{his} \textcolor{orange}{own} \textcolor{orange}{pages} \textcolor{blue}{with} \textcolor{blue}{dece} \textcolor{blue}{it} \textcolor{orange}{.} \textcolor{orange}{May} \textcolor{orange}{h} \textcolor{blue}{ap} \textcolor{blue}{a} \textcolor{blue}{s} \textcolor{orange}{ly} \textcolor{orange}{ly} \textcolor{orange}{.} \textcolor{orange}{p} \textcolor{blue}{enn} \textcolor{orange}{ies} \textcolor{orange}{,} \textcolor{orange}{a} \textcolor{orange}{d} \textcolor{orange}{agger} \textcolor{blue}{ty} \textcolor{orange}{,} \textcolor{orange}{whispered} \textcolor{blue}{threat} \textcolor{blue}{sends} the \textcolor{blue}{w} \textcolor{orange}{iser} \textcolor{orange}{f} \textcolor{orange}{'} \textcolor{orange}{r} \textcolor{orange}{their} \textcolor{orange}{lives} \textcolor{orange}{,} \textcolor{blue}{yet} \textcolor{blue}{in} \textcolor{blue}{the} \textcolor{orange}{end} \textcolor{orange}{,} \textcolor{orange}{might} \textcolor{blue}{t} \textcolor{blue}{is} \textcolor{blue}{only} \textcolor{orange}{f} \textcolor{orange}{'} \textcolor{orange}{r} \textcolor{orange}{a} \textcolor{orange}{tragic} hero \textcolor{blue}{like} \textcolor{blue}{Sean} Be \textcolor{blue}{an} \textcolor{blue}{to} \textcolor{orange}{pay} \textcolor{orange}{the} \textcolor{blue}{ultimate} price \textcolor{blue}{.} \textcolor{orange}{W} \textcolor{orange}{oe} \textcolor{blue}{bet} \textcolor{blue}{ide} \textcolor{blue}{ye} \textcolor{blue}{,} \textcolor{orange}{ye} \textcolor{orange}{tre} \textcolor{blue}{acher} \textcolor{blue}{ous} \textcolor{orange}{qu} \textcolor{blue}{ill} \textcolor{blue}{.} \textcolor{orange}{</s>} \\

\end{tabular}
\caption{Examples of dooley-scaled generations that approach optimality. Green tokens are unigram scaled, blue bigram, and green trigram.}
\label{tab:dooleyfull}
\end{table*}

Table~\ref{tab:dooleyfull} collects a series of roughly optimal dooley-scaled generations. Each displays a unique combination of transferred features, and helps demonstrate the smooth nature of this method of transfer.



\end{document}